\documentclass[runningheads]{llncs}
\usepackage{graphicx}
\usepackage{amsmath,amssymb} 
\usepackage{color}
\usepackage{subfig} 
\usepackage[width=122mm,left=12mm,paperwidth=146mm,height=193mm,top=12mm,paperheight=217mm]{geometry}

\newcommand{\tld}{\raise.17ex\hbox{$\scriptstyle\mathtt{\sim}$}}

\begin{document}
\pagestyle{headings}
\mainmatter

\title{DenseNet: Implementing Efficient ConvNet Descriptor Pyramids}
\subtitle{Technical Report}
\titlerunning{DenseNet}
\authorrunning{Iandola, Moskewicz, Karayev, Girshick, Darrell, and Keutzer}
\author{Forrest Iandola, Matt Moskewicz, Sergey Karayev, \\
		Ross Girshick, Trevor Darrell and Kurt Keutzer}
\institute{University of California, Berkeley}
\maketitle

\begin{abstract}
Convolutional Neural Networks (CNNs) can provide accurate object classification. They can be
extended to perform object detection by iterating over dense or selected proposed object
regions. However, the runtime of such detectors scales as the total number and/or area of regions to
examine per image, and training such detectors may be prohibitively slow. However, for some CNN
classifier topologies, it is possible to share significant work among overlapping regions to be
classified. This paper presents DenseNet, an open source system that computes dense, multiscale
features from the convolutional layers of a CNN based object classifier. 
Future work will involve training efficient object detectors with DenseNet feature descriptors.

\end{abstract}

\begin{figure}
\centering
\fbox{
    \includegraphics[width=12cm]{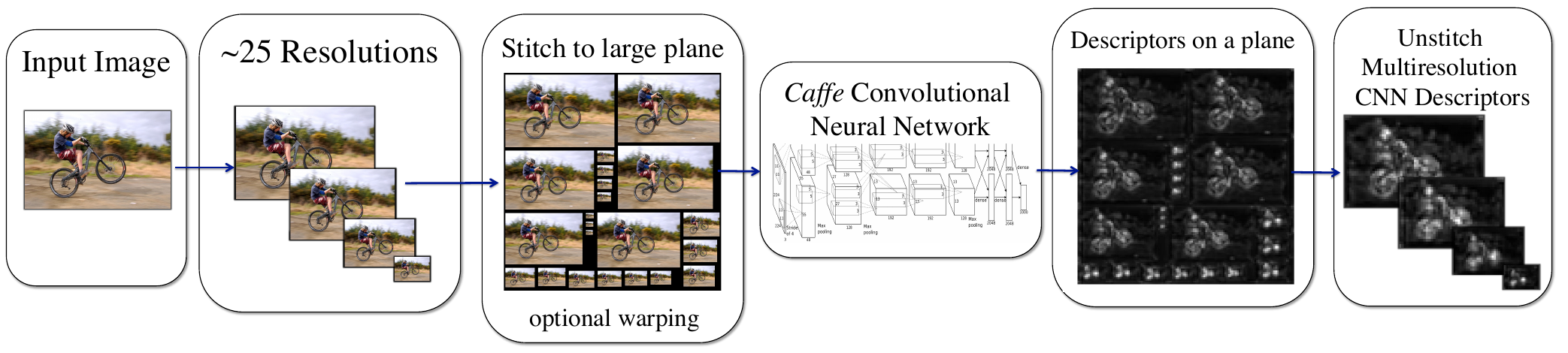}
}
\caption{DenseNet multiscale feature pyramid extraction}
\label{fig:DenseNet}
\end{figure}

\section{Introduction}
\label{sec:Intro}
The rebirth of deep neural networks has led to profound improvements in the accuracy of object recognition algorithms.
The key algorithms of the deep learning revolution can be traced back to the late 1980s~\cite{LeCun-1989}.
However, the rise of big data has led to huge labeled datasets (e.g. ImageNet~\cite{imagenet_cvpr09} with $>$1M labeled images) for training and evaluating object recognition systems.
Additionally, extremely efficient deep neural network implementations such as Berkeley's {\em Caffe}~\cite{caffe-code} and Toronto's {\em cuda-convnet}~\cite{Alexnet}  expose enough parallelism to make ImageNet a tractable benchmark for deep neural network object classification. The Caffe framework is also designed to encourage research, development, and collaboration though a robust open source development model and a rich set of features for configuration, testing, training, and general CNN experimentation.
Today, deep convolutional neural networks (CNNs) such as Alexnet~\cite{Alexnet} and Clarifai~\cite{Clarifai} produce state-of-the-art object classification accuracy (up to 88\% when scored on top-5 categories) on the 1000-category ImageNet dataset. Further, in areas such fine-grained recognition and image segmentation, using ImageNet-trained deep CNNs as feature descriptors has advanced the state-of-the-art accuracy substantially~\cite{DeCAF}. 

Perhaps the biggest success story for cross-domain portability of ImageNet-trained CNNs lies in
object detection.  While {\em object classification} problems provide labels or crops to indicate
object locations, {\em object detection} requires both object localization and classification.  

\begin{figure}
\centering
\fbox{
    \includegraphics[width=11cm]{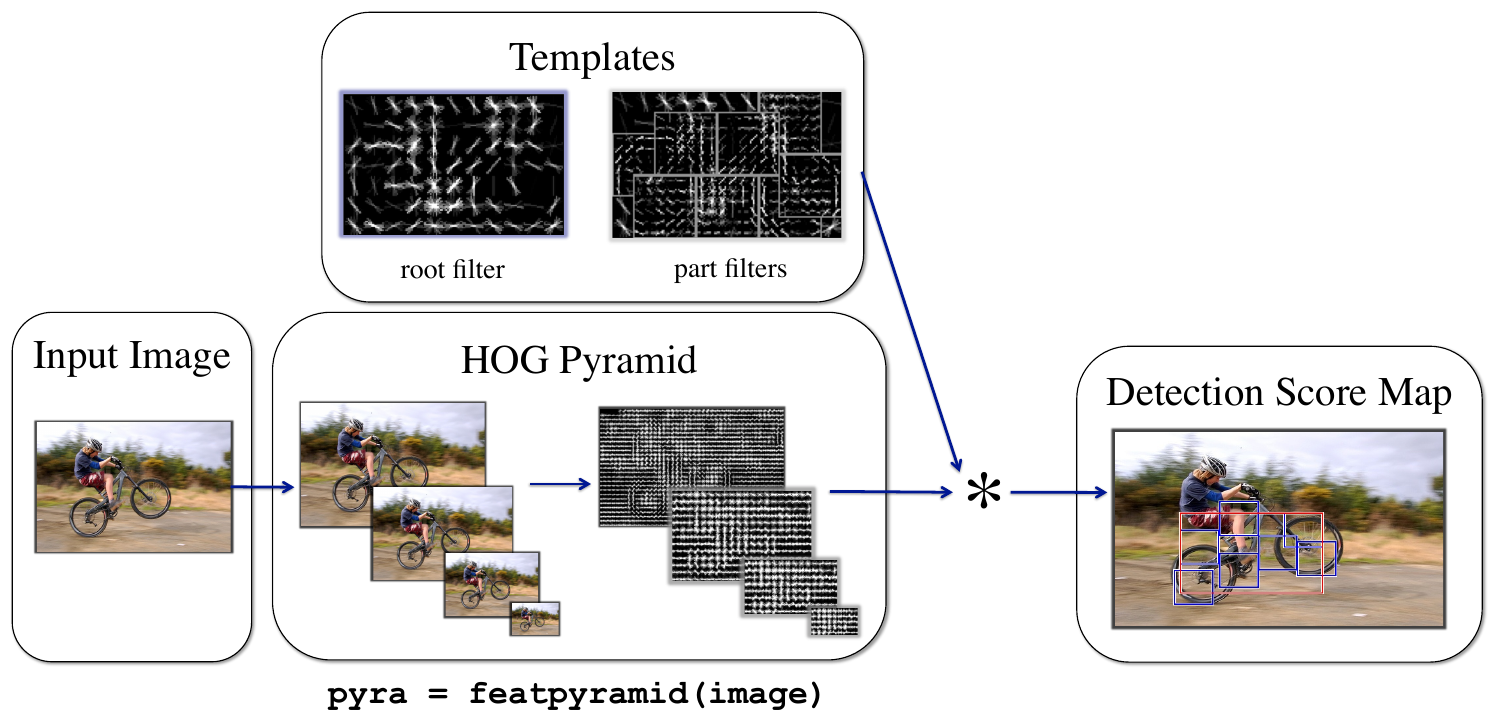}
}
\caption{Sliding-window object recognition. A number of detection methods including Deformable Parts Models (DPMs) and Poselets are built on pipelines like this.}
\label{fig:DPM}
\end{figure}

The sliding window approach is common way to convert a object classifier into an object
detector. Compared to sparse approaches, one key property is that it is trivial to create the set of
region proposals. However, depending on the desired density in position and scale, the number of
region proposals may become quite large. Traditionally, algorithms such as Deformable Parts Models
(DPMs)~\cite{FelzenszwalbPAMI10} and Poselets~\cite{Poselets} achieved high-quality object detection
with sliding-window multi-template detectors on HOG~\cite{HOG} features.  The best of the
sliding-window detector breed have typically yielded around 33\% mean average precision (mAP) on the
PASCAL~\cite{PASCAL} 2007 object detection dataset; using additional hand-engineered descriptors can
yield incremental improvements.

Extending a current state of the art CNN-based classifier into an object detector using a naive
dense sliding window set of region proposals would be prohibitively slow. For example, with a
per-region classification time of \tld50ms and \tld200K regions, the per image detection time would
be \tld3 hours. To avoid this issue, one can either reduce the number of region proposals, decrease
the time spent per region, or some combination of the two. In particular, CNNs offer the potential
to share significant work between overlapping regions. In the related work section, we consider
various approaches that explore specific design points in this space of options.

The remainder of the paper is organized as follows: In Section~\ref{sec:Related}, we review related
work on dense CNN features, particularly for object detection. Section~\ref{sec:DenseNet} proposes
{\em DenseNet}, our approach to efficiently extracting pyramids of CNN descriptors. We discuss using
the DenseNet features to efficiently support the classification method of~\cite{Alexnet} over many
possible region proposals for an image; in particular, we discuss the key issues of supporting
per-region data centering, varied sizes, and varied aspect ratios.

\section{Related Work}
\label{sec:Related}

\noindent
{\bf CNNs for Object Detection.}

DetectorNet~\cite{DetectorNet} performs sliding-window detection on a coarse 3-scale CNN
pyramid.  Due to the large receptive field of CNN descriptors, localization can be a challenge for
sliding-window detection based on CNNs.  Toward rectifying this, DetectorNet adds a procedure to
refine the CNN for better localization.  However, DetectorNet does not pretrain its CNN on a large
data corpus such as ImageNet, and this may be a limiting factor in its accuracy.
In fact, DetectorNet reported 30\% mAP on PASCAL VOC 2007, less than the best HOG-based methods.

OverFeat~\cite{OverFeat} generates dense, multi-scale CNN features suitable for object detection and
classification for square regions. OverFeat does not consider the issue of non-square region
proposals as they are not necessary for their approaches to detection or classification. In our
approach, we support both the extraction of non-square regions of feature descriptors as well as the
higher level approach of constructing multiple feature pyramids where, for each pyramid, the input
has been warped to a selected aspect ratio. While precompiled binaries for running the OverFeat CNN to create
such features using provided pre-trained CNN model parameters are provided, training code is
explicitly not provided. Further, it is unclear how much of the source code for the rest of the
OverFeat system is available; some is part of the Torch7~\cite{Torch7} framework used by
OverFeat. This lack of openness hinders the usage of OverFeat as the basis for coupled CNN /
detection algorithm design space exploration and comparison, particularly for benchmark sets where
no OverFeat-based detection results are available (such as the PASCAL objection detection
benchmarks). Also, unlike the Caffe system into which DenseNet is integrated, it appears that
OverFeat does not focus on providing a robust, general, open platform for research, development, and
efficient GPU computation of CNNs.



Another recent approach called Regions with Convolutional Neural Network features (R-CNN)~\cite{R-CNN}
leverages an ImageNet-trained CNN as a feature descriptor to achieve a profound boost in accuracy:
54\% mAP on PASCAL 2007, and up to 59\% with bounding box regression.
Unlike traditional sliding-window detection approaches, R-CNN decouples
the localization and classification portions of the object detection task.  R-CNN begins by
generating class-independent region proposals with an algorithm such as Selective Search~\cite{SS}.
Then, it extracts CNN descriptors on the proposed regions after warping them to a fixed square size.
Finally, R-CNN scores and classifies the proposed regions using a linear SVM template on the CNN
descriptors.


\begin{figure}
\centering
\fbox{
    \includegraphics[height=3.5cm]{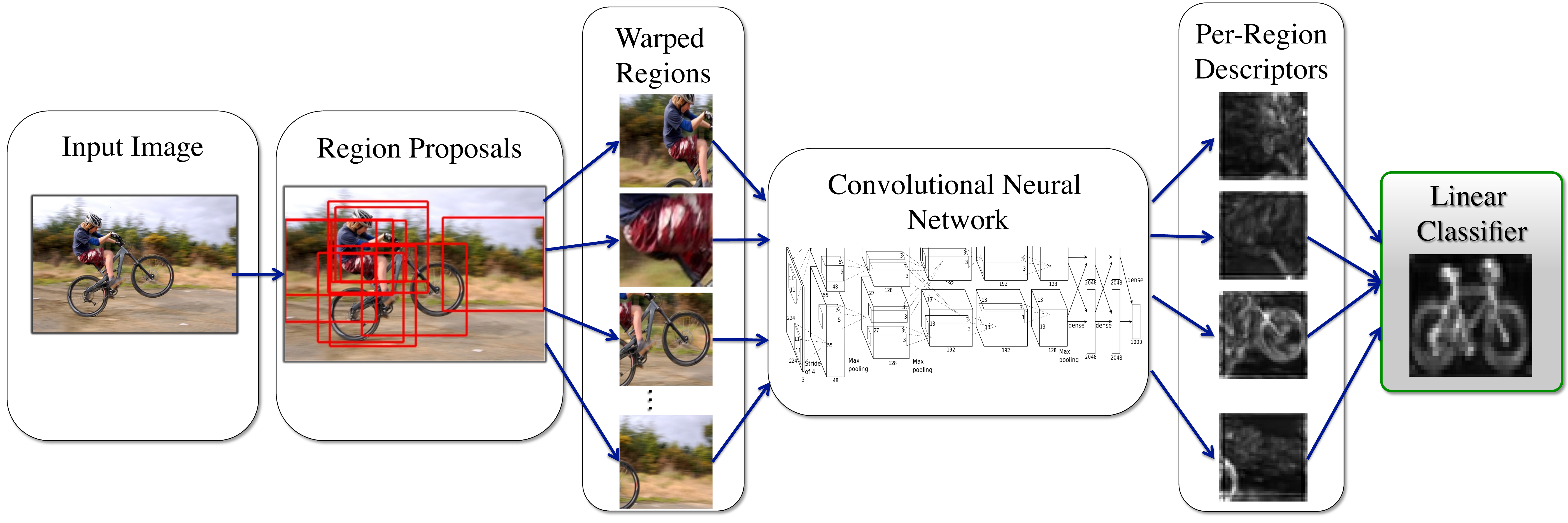}
}
\caption{Object Detection with R-CNN~\cite{R-CNN}: region proposals and CNN descriptors.}
\label{fig:RCNN}
\end{figure}

Currently, the overall runtime of R-CNN yields a latency of \tld10s per image. This latency renders
the approach unsuitable for interactive applications such as image labeling or search. However,
since many of the region proposals for a given image overlap, much image area is being processed by
the CNN many times. Further, the bulk of the computation of the CNNs occurs in its early
convolutional layers and does not depend on the relative position of image patches within
regions. This suggests that it may be possible to share a great deal of work among all the region
proposals for a given image. However, data centering issues and the fact that the regions are of
differing sizes and aspect ratios makes this reuse more difficult to achieve. Using DenseNet to
achieve such reuse for the R-CNN algorithm without significant loss of accuracy is a subject of
ongoing work. 
{\bf Other Uses of Dense and Multiscale CNN Pyramids.} \\

A number of approaches have arisen for computing dense pyramids of CNN features in various computer vision applications outside of object detection.
Farabet et al.~\cite{farabet-scene-parsing} and Jiu et al.~\cite{jiu-human-estimation} construct multiresolution pyramids of 2-layer CNNs.
Farabet et al. apply their network to scene parsing, and Jiu et al. showed multiscale CNN results on human pose estimation.
Along the same lines, Giusti et al. compute CNN pyramids and perform sliding-window processing for image segmentation~\cite{Giusti:13}.
Several years earlier, Vaillant et al. densely computed CNNs on full images for robust face localization~\cite{Vaillant:94}.

\section{DenseNet CNN Feature Pyramids}
\label{sec:DenseNet}
The CNN object classifier of~\cite{Alexnet} operates on fixed size square images. The bulk of the
computation performed by the classifier occurs in the first five convolutional layers of the neural
network and takes time roughly proportional to the number of input pixels. As a preprocessing step,
the input image is centered by subtracting the mean image created from a large data set. However,
after this step, the computation performed in the convolutional layers is translationally
invariant. Thus, the output of a node at the final convolution layer depends only on the value of
the supporting image patch, not on the relative location of the node within its image plane. Hence,
for two overlapping region proposals of the same size and aspect ratio, the values of any nodes that
share the same supporting image patch will be identical, and need not be recomputed. Consider the
following simplified example: assume a classifier that operates on images of size \(M*M\) (ex:
\(200\)) and an input image of size \(N*N\) (ex: \(1000\)). With a stride of 16px, there are
\(\tld{}R=((N-M)/16)^2\) (ex: \(\tld2.5K\)) possible \(M*M\) regions within the \(N*N\)
region. Computing the convolutional layers on these \(R\) regions takes time proportional to
\(R*M*M\) (ex: \(\tld100Mops\)). However, computing the convolutional layers on the original image
directly only takes time proportional to \(1*N*N\) (ex: \(\tld1M ops\)). Thus, a single full-image
dense computation of the features yields a speedup of \(100X\) over computing the features
per-region.

However, we wish to accelerate classification over broader set of aspect ratios and sizes of regions
within the input image, and we must somehow deal with the mean-image data centering issue as
well. We will detail our approach to these problems in the following sections. In summary:
\begin{itemize}
\item For the issue of differing scales, we take the traditional approach of construction a multi-resolution pyramid of images formed by up- and down- sampling the input image with a configurable selection of scales. Additionally, we deal with some complexities of processing such pyramids in the {\em Caffe}~\cite{caffe-code} system.
\item For the issue of data centering, we choose to center using a single mean pixel value, rather than a mean image, and provide some experimental justification that this simplification is acceptable.
\item For the issue of multiple aspect ratios, we choose to push the problem downstream to later detector stages, but also consider the possibility of creating multiple image pyramids at a selection of aspect ratios. 
\end{itemize}


\subsection{Multiscale Image Pyramids for CNNs}

We show the overall flow of our DenseNet multiscale feature extractor in Figure~\ref{fig:DenseNet}. The selection of scales chosen for our pyramids is similar to that used for other features in other object detectors, such as the HOG feature pyramids used by the DPM object detectors of~\cite{release5}. The maximum scale is typically 2, and the minimum scale is chosen such that the down-sampled image maintains a particular minimum size (often \tld16-100px). There are typically 3, 5, or 10 scales per octave (depending on application), yielding pyramids with \tld10-50 levels and \tld3-8X the total number of pixels of the original image. 

A key factor in the rebirth of CNNs is the rise of efficient CPU and GPU implementations such as cuda-convnet~\cite{Alexnet}, Torch 7~\cite{Torch7}, and Caffe~\cite{caffe-code}. To maximize computational efficiency, these CNN implementations are optimized for {\em batch mode}, where tens or hundreds of equal-sized images are computed concurrently. However, to compute feature pyramids, our goal is to compute CNN descriptors on an input image sampled at many resolutions. Thus, our multi-resolution strategy is at odds with the CNN implementations' need for batches of same-sized images.

However, with at least the Caffe implementation of CNNs, a single large image (with a similar total pixel count as a normal batch) can also be efficiently computed. Using the Bottom-Left Fill (BLF) algorithm implemented in~\cite{FFLD}, we stitch the multiple scales of the input image pyramid onto as many large (often of size 1200x1200 or 2000x2000, depending on available GPU memory) images as needed, and then run each individual image as a batch. Finally, we unpack the resulting stitched convolutional feature planes into a feature pyramid.

Using this approach, however, in turn raises another issue. Given the kernel/window sizes of the convolutional and max-pooling commonly found in CNNs, each descriptor from a deep convolutional layer can have a large (perhaps \tld200px) receptive field size (or supporting image patch size) in the input image. Thus, stitching could lead to edge/corner artifacts and receptive field pollution between neighboring pyramid scales that are adjacent in the large stitched images. To mitigate this, we add a 16px border to each image, for a total of at least 32px of of padding between any pair of images on a plane. We fill the background with the mean pixel value used for data centering (as discussed below). Finally, we linearly interpolate all image padding between the image's edge pixel and the centering mean pixel value. Experimentally, we find that this scheme seems successful in avoiding obvious edge/corner artifacts and receptive field pollution.


\subsection{Data Centering / Simplified RGB mean subtraction}
The CNN classifier of~\cite{Alexnet} subtracts a mean image (derived from ImageNet) from each input image to center it prior to feeding it into the CNN. For stitched images containing many pyramid levels, or even for a single image that is to be processed to support many possible region proposals, it is unclear how achieve per-region centering by a mean image. Therefore, we instead use a mean {\em pixel} value (derived from the ImageNet mean image). 

Remember from the previous subsection that we fill the background pixels in our planes with the mean ImageNet mean RGB value, so the background pixels on planes end up being zeros after centering.

As a validation of this simplified mean pixel centering scheme, we run a pretrained Alexnet model on ImageNet. 
Running with one RGB mean pixel is 0.2\% less accurate than using the RGB mean mask for top-1 classification.
Thus, our simplification of using a mean pixel does not appear to substantially affect accuracy.

\subsection{Aspect Ratios}
For the most part, we choose to delegate the handling of different aspect ratios to later stages in the detection pipeline. In particular, such stages may utilize multiple templates with various aspect ratios or warp regions in feature space using non-square down-sampling such as non-square max-pooling. However, note that for any selection of interesting aspect ratios it is possible to, for each aspect ratio, warp the input image and construct an entire warped feature pyramid as per the above procedure. Of course, the overall feature computation time scales as the number of desired aspect ratios. 


\subsection{Measured Speedup}

We observe that it takes 10sec to compute 2000 window proposals in traditional Alexnet network in Caffe.
DenseNet takes 1sec to compute a 25-scale feature pyramid.
We conducted these experiments on an NVIDIA K20 GPU.


\noindent
\subsection{Straightforward Programming Interface}
We provide DenseNet pyramid extraction APIs for Matlab and Python integrated into the open source Caffe framework.
Our API semantics are designed for easy interoperability with the extremely popular HOG implementation in the Deformable Parts Model (DPM) codebase: \\

\noindent
DPM HOG: {\tt pyra = featpyramid(image)} \\
DenseNet: {\tt pyra = convnet\_featpyramid(image\_filename)}

\section{Qualitative Evaluation}
One of our main goals in densely computing CNN descriptors is to avoid the computational cost of independently extracting CNN descriptors for overlapping image regions.
Thus, it is important that we evaluate whether or not our dense CNN descriptors can approximate CNN descriptors that are computed for individual image regions in isolation.
In other words, when computing descriptors for regions of an image, how different do the descriptors look whether we crop the regions \emph{before} or \emph{after} doing the CNN descriptor computation?

\begin{figure}[p]
\centering
\fbox{
    \includegraphics[height=4.5cm]{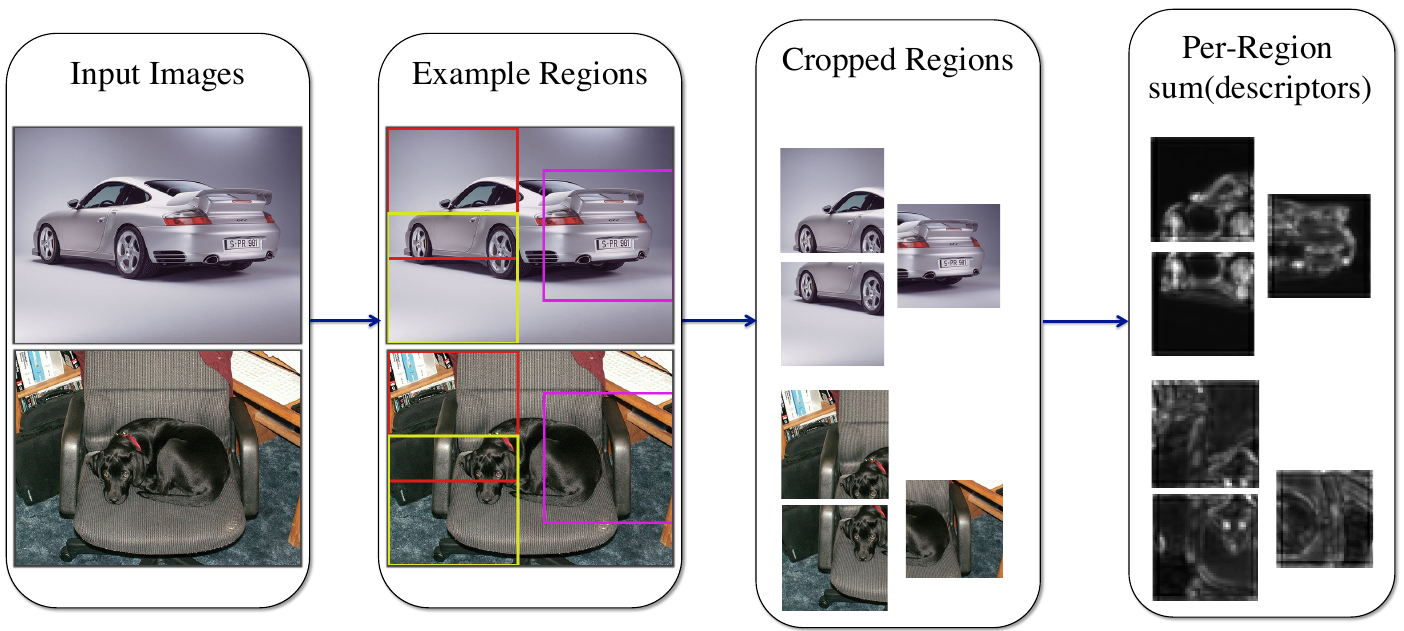}
}
\caption{Descriptors independently computed on image regions. Here, we first crop regions from images, then compute descriptors. This is the type of approach used in R-CNN~\cite{R-CNN}. The image regions were chosen arbitrarily, not taken from~\cite{SS}. Also notice that the image regions used in this example are square, so no pixel warping is needed.}
\label{fig:cropPixels}
\end{figure}

\begin{figure}
\centering
\fbox{
    \includegraphics[height=4.5cm]{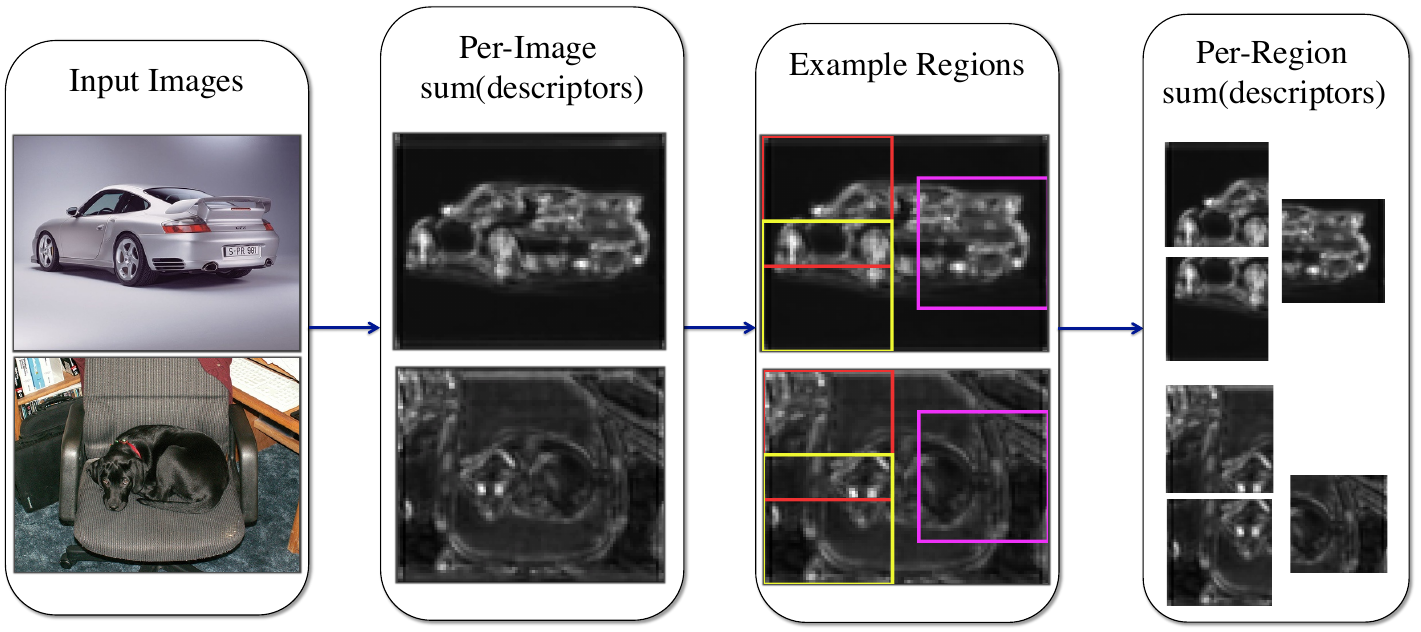}
}
\caption{Descriptors computed on a full image. These descriptors can be cropped to approximate region proposal windows (rightmost panel). DenseNet is optimized for this type of feature extraction.}
\label{fig:cropDescriptors}
\end{figure}

To perform this evaluation, we visualize some example images in two different feature extraction pipelines.
In Figure~\ref{fig:cropPixels}, we crop regions from pixel space and compute descriptors on each window independently. 
This pipeline is computationally inefficient with large numbers of regions, but this pipeline serves as our baseline for descriptor computation in applications such as R-CNN~\cite{R-CNN}.
In contrast to Figure~\ref{fig:cropPixels}, DenseNet first computes descriptors densely without regard for region proposal windows, and then regions can be cropped from DenseNet descriptor pyramids. 
In Figure~\ref{fig:cropDescriptors}, we show an example scale from a DenseNet descriptor pyramid, and we crop descriptor regions based on the same regions used in Figure~\ref{fig:cropPixels}.
Note that descriptors are visualized as the sum over channels -- this preserves spatial resolution while sufficiently reducing dimensionality for straightforward 2D visualization.
The key takeaway here is that the descriptors in the rightmost boxes of Figures~\ref{fig:cropPixels} and~\ref{fig:cropDescriptors} look similar, so DenseNet appears to be a good approximation of per-region descriptors computed in isolation.


\section{Conclusion}
\label{sec:Conclusion}
We have presented an overview of how we densely compute CNN feature pyramids with DenseNet.

\clearpage

\bibliographystyle{splncs}
\bibliography{paper}
\end{document}